\documentclass[final]{cvpr}

\usepackage{times}
\usepackage{epsfig}
\usepackage{graphicx}
\usepackage{amsmath}
\usepackage{amssymb}
\usepackage{algorithm}
\usepackage{algpseudocode}

\usepackage[pagebackref=true,breaklinks=true,colorlinks,bookmarks=false]{hyperref}

\def\cvprPaperID{15} 
\def\confYear{CVPR 2021}

\begin{document}

\title{SegVisRL: Visuomotor Development for a Lunar Rover for Hazard Avoidance using Camera Images}

\author{Tamir Blum,


Gabin Paillet,

Watcharawut Masawat,

Mickael Laine
and
Kazuya Yoshida\\

{\tt\small tamir@dc.tohoku.ac.jp}\\

Department of Aerospace Engineering, Tohoku University\\
Aoba 6-6-01, Aramaki, Aoba-ku, Sendai, Miyagi
980-8579, Japan\\
}

\maketitle

\begin{abstract}
  The visuomotor system of any animal is critical for its survival, and the development of a complex one within humans is large factor in our success as a species on Earth. This system is an essential part of our ability to adapt to our environment. We use this system continuously throughout the day, when picking something up, or walking around while avoiding bumping into objects. Equipping robots with such capabilities will help produce more intelligent locomotion with the ability to more easily understand their surroundings and to move safely. In particular, such capabilities are desirable for traversing the lunar surface, as it is full of hazardous obstacles, such as rocks. These obstacles need to be identified and avoided in real time. This paper seeks to demonstrate the development of a visuomotor system within a robot for navigation and obstacle avoidance, with complex rock shaped objects representing hazards. Our approach uses deep reinforcement learning with only image data. In this paper, we compare the results from several neural network architectures and a preprocessing methodology which includes producing a segmented image and downsampling. 
\end{abstract}

\section{Introduction}

The ability to understand the environment and adapt locomotion to it is a basic capability often found in intelligent lifeforms across animals. It is used by humans in our daily lives, whether picking something up, walking outside and avoiding to step on sharp rocks or navigating inside a building without bumping into clutter. The desire to expand this capability to robots has been one of the driving forces behind Reinforcement Learning (RL) research applied to robotics. 

Our daily lives consist mostly in a man-made environment, filled with many hurdles (such as stairs, sidewalks) and obstacles (such as furniture, other people). We take these factors into account during locomotion making sure not to hit anything or stumble. Likewise, the natural world is also filled with many such hurdles (terrain differences, slopes) and obstacles (trees, rocks). 

\begin{figure}[t]
\begin{center}
   \includegraphics[width=1\linewidth]{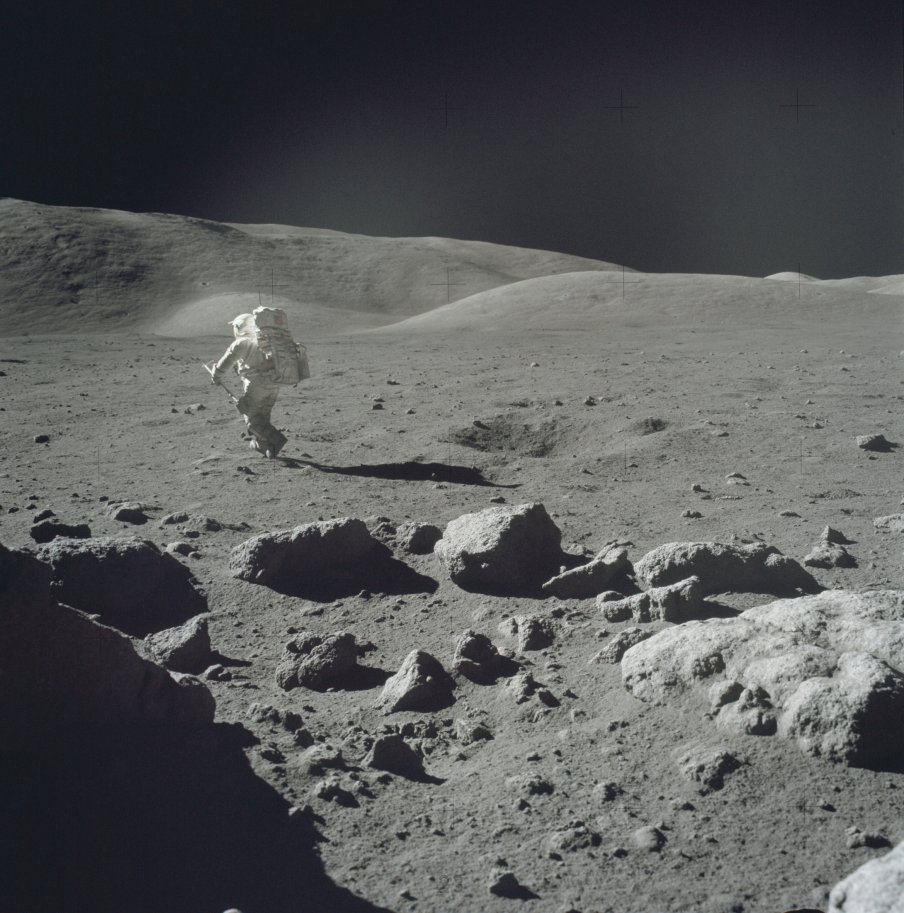}
\end{center}
   \caption{A photo of the lunar surface taken during an Apollo mission. In this photo, large rocks are which pose a danger to a rover traversing on the lunar surface can be seen. Such rocks would not be visible on satellite images. Credit: NASA}
\label{fig:lunarsurface}
\end{figure}

In celestial bodies, many of the environments are extreme in terms of bumpy ground or variable friction surfaces and are filled with hazards such as rocks of different sizes and craters. As such, much research has been conducted on how to create safe and efficient locomotion, path planning and obstacle avoidance for space exploration robots as well\cite{ono_maars_2020}. 

Satellite images exist which can be used to help provide long-distance path planning. However, resolution from such satellites is insufficient to identify all hazards. For example, the Lunar Reconnaissance Orbiter has produced a global map of the moon with resolution of around 100 m/pixel. Thus, rovers need to be able to adapt to their environment in realtime to avoid the small-medium hazards on the lunar surface, such as those shown in Figure \ref{fig:lunarsurface}.

This paper seeks to combine the RL and computer vision and to demonstrate the application of Reinforcement Learning to develop a visuomotor system within a robot using visual data alone. This data is preprocessed and includes information about the goal location, and obstacle locations within the robots field of view. 

In particular, we demonstrate the use of RL for a partially observable system, that is, a system in which the learning agent does not have access to perfect and all-encompassing state information of the environment (i.e. we do not have access to the velocity of the rover, the angular rotation or the position directly). We show that through the use of Deep Reinforcement Learning (DRL) with neural networks (NN) as the function approximator for the agent, or decision making system, that a robot can solve such problems.

\begin{figure}[b]
\begin{center}
   \includegraphics[width=1\linewidth]{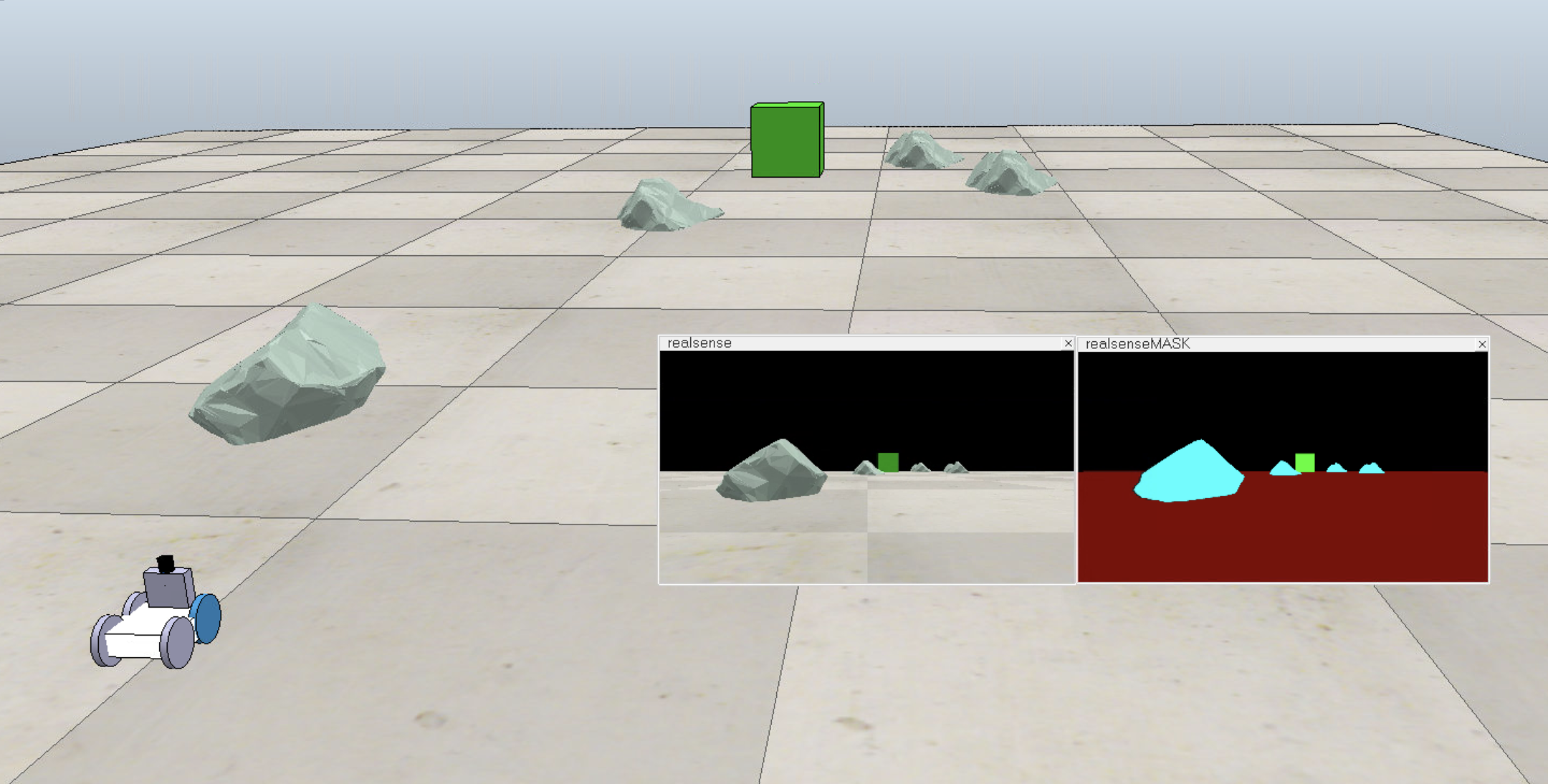}
\end{center}
   \caption{A small rover learns to navigate in a world with complex rock shapes that pose a hazard to the rover and must be identified and avoided in real time.}
\label{fig:clover}
\end{figure}
\begin{figure}[b]
\begin{center}
   \includegraphics[width=1\linewidth]{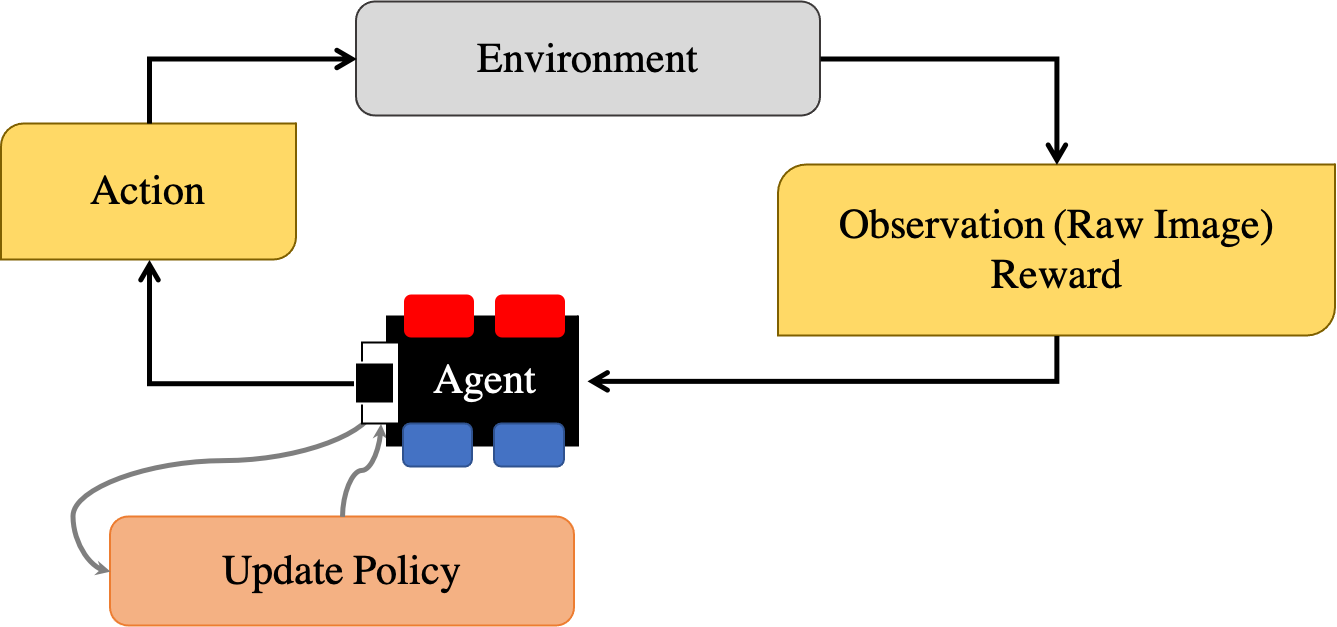}
\end{center}
   \caption{The agent learns by interacting with the environment, with access only to the reward and observation. This is a case of a Partially Observable Markov Decision Process.}
\label{fig:rlsystem}
\end{figure}

\section{Background}
\subsection{Reinforcement Learning}
The goal of the RL algorithm is to train an agent through optimisation to a reward function to solve a Markov Decision Process (MDP). A MDP describes a stochastic problem where a single action given a certain state could result in a number of resulting subsequent states. The transition probability $P$ describes the probability to transition from said state to said subsequent state, and is given by the probability that the transition occurs given the proceeding state, $s$ from within the set of possible states $S$, and a certain action, $a$ from within the set of possible actions $A$. 

\begin{equation}
  P(s, s') = Pr(S_{t+1}=s'|S_t=s,A_t=a)
\end{equation}

The reward function is tuned to the problem by the engineer, in a process known as reward shaping, and is what allows the agent to differentiate good actions from bad ones. In order to deter the agent from making short term decisions that could harm itself in the long term, we take into account not just the current reward but the expectation of future rewards. Since expectations contain uncertainty and volatility, we discount these terms exponentially w.r.t time. 

\begin{equation}
  G_t = R_{t+1} + \gamma^{1}R_{t+2} + \gamma^{2}R_{t+3} + … + \gamma^{n}R_{t+n} 
\end{equation}
\begin{equation}
\nonumber
  G_t  = \sum_{k=0}^\infty \gamma^{k}R_{t+k+1}
\end{equation}
The only term not to contain uncertainty is $R_{t+1}$, which describes the reward given the action $a$ at time $t$.  

A key assumption within MDPs is independence of the future w.r.t the past. 
\begin{equation}
  P(S_{t+1}|S_t) = P(S_{t+1}|S_1,...,S_t)
\end{equation}

The agent can hold a policy, $\pi$ which is contained within a neural network for DRL. In stochastic algorithms, this policy would be a probability distribution of actions over states.

\begin{equation}
    \pi(a|s) = Pr(A_t=a|S_t=s) 
\end{equation}

\subsubsection{PPO}
This paper uses PPO\cite{schulman_proximal_2017}, a popular RL algorithm, self described as being simpler to tune with good performance compared to similar algorithms. One of the key components of this algorithm is the clipping function it uses for the surrogate object function, $L$. This algorithm often uses Stochastic Gradient Decent to update the policy function approximator, i.e. the neural network. 

\begin{equation}
  L^{CLIP}(\theta) = 
\end{equation}
\begin{equation}
\nonumber
  \hat{E}_t[min(r_t(\theta)\hat{A}_t,clip(r_t(\theta),1-\epsilon, 1+\epsilon)\hat{A}_t)]
\end{equation}

\begin{figure}
\begin{algorithm}[H]
\caption{PPO Algorithm}
\label{ppoalg}
\begin{algorithmic}
 \State initialize Neural Network, RL algorithm parameters\;
 \While{$Training = True$}{
   \State Run policy $\pi_{\theta_{old}}$ in environment for $T$ timesteps\;
   \State Compute advantage estimates $\hat{A}_1,...,\hat{A}_T$\;
   \State Optimize surrogate $L(\theta)$ with $K$ epochs and Minibatch size $M \leq N_{step}$\;
   \State $\theta_{old} \xleftarrow{} \theta$\;
   \EndWhile
 }
\end{algorithmic}
\end{algorithm}
\end{figure}

\section{Related Works}
In this section we discuss related works to AI for robotics and space robotics applications. 

\subsection{Reinforcement Learning for Robotics}
Reinforcement learning has been applied to robotics in a variety of settings. 

These cases include applying DRL with traditional techniques such as SLAM\cite{zhang_neural_2020} for exploring unknown environments. It has also been used with robot state information to learn path planning and motion control \cite{blum_ppmc_2020}\cite{blum_ppmc_2020-1} for both walking robots and rovers, as well as for mixing robot state information with lidar information to avoid colliding with walls\cite{tai_virtual--real_2017}. 

There has also been some work done directly on the robot itself rather than through simulation \cite{haarnoja_learning_2019}, while other works train in simulation and then transfer it over in a process known as simulation to reality transfer (sim2real)\cite{tai_virtual--real_2017}. Another  focus area is primarily on the simulation itself\cite{blum_rl_2020}. Lidar-like terrain depth has also been used to identify terrain and respond accordingly in a 2-D environment \cite{peng_terrain-adaptive_2016}.

While in this paper we use end-to-end control, meaning that the RL goes from raw data to direct outputs. An alternative is to use more modular approaches such as hierarchical control\cite{peng_deeploco_2017} or feeding the RL into a controller. For robotics applications, many variables need to be taken into account, such as the action action space\cite{peng_deeploco_2017}.

A technique called imitation learning to copy animal movements \cite{bin_peng_learning_2020} has also been widely researched. 

\subsection{Space Robotics}
AI for space applications has been of interest to researchers due to the high requirement for autonomy and optimization in a resource limited environment. This is due both to extreme conditions and transmission delay/limitations. 

This includes but is not limited to reinforcement learning used for controls \cite{surovik_adaptive_2018}\cite{bonardi_novel_2020}\cite{sakamoto_evaluation_2020}.

Non-reinforcement learning cases are numerous, such as anomaly detection to try to detect areas of high interest via unsupervised learning \cite{stefanuk_detecting_2020}, or classifying terrain via supervised learning\cite{raimalwala_enabling_2020}. AI has also been used to help with slip prediction for avoiding entrapment\cite{inotsume_slip_2020}.

\section{Visuomotor System Development: Myopic Path Planning from Pixels to Actions}
This section describes how we use DRL to develop the visuomotor system for a rover in simulation to conduct path planning, motion control and obstacle avoidance. 

\subsection{Training Procedure and Training Algorithm}
A customized training algorithm was used alongside a preprocessing methodology and DRL. The algorithm is shown in Algorithm \ref{ppmcalg}.

\begin{figure*}
\begin{center}
\includegraphics[width=\textwidth,height=16cm]{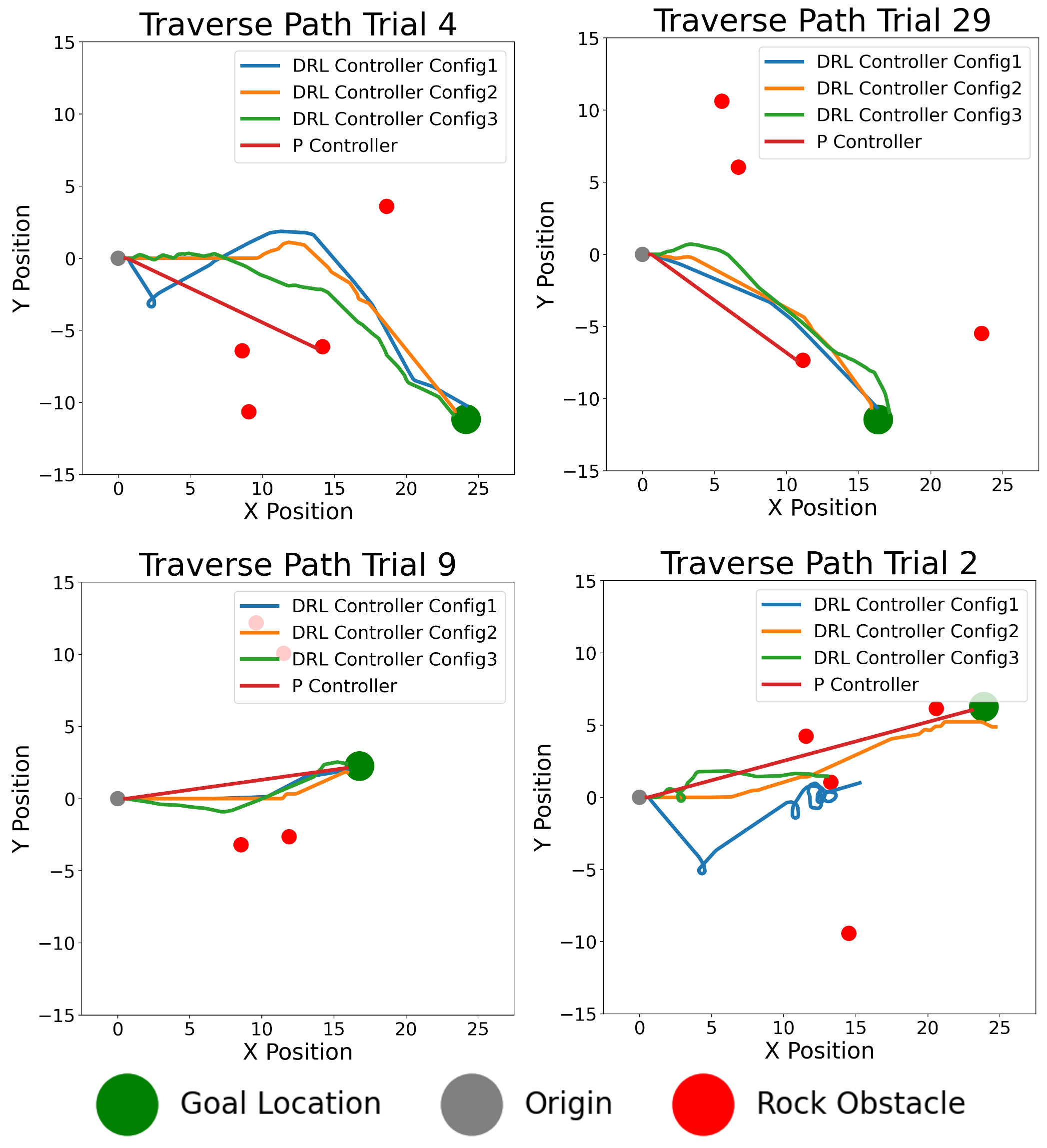}\hspace{0pt}

\caption{Four of the 30 trials are shown. The top two figures show cases where all the DRL controllers were able to reach the goal, while the P controller crashed into an obstacle. The bottom left figure shows a case where all controllers reach the goal. Lastly, the bottom right case shows an instance where the DRL Controller Config 1 was able to dodge the obstacle, however, ran out of time and failed the episode.}
\label{fig:Traverse}
\end{center}
\end{figure*}

A 25m x 25m map was created with the rover starting at the middle of one end of the map, as shown in Figure \ref{fig:Traverse}. The goal of the rover is to move from the origin to the goal, while avoiding obstacles, within the time limit. 

During training, a collision was recorded if the rover got within $0.5m$ of the center of any of the obstacles, whereas a win was recorded if the rover got within $1m$ of the center of the goal. The goal location and $4$ obstacle locations were randomly chosen at the start of each episode. The goal was placed at least $10 meters$ away from the starting point of the rover. The obstacles were places at least $4 meters$ away from the starting point of the rover.

\begin{figure}
\begin{algorithm}[H]
\caption{Training Algorithm}
\label{ppmcalg}
\begin{algorithmic}
 \State initialize Neural Network, RL algorithm\;
 \While{Training = True}
  \State Step simulation forward
  \If{Time $<$ Max Time}
   \State Feed state to policy, output action to robot
   \State Calculate reward
   \If{Endpoint Reached}
    \State Reset Episode = True
   \ElsIf{Fail Condition Reached}
    \State Reset Episode = True
   \EndIf
  \Else
   \State Reset Episode = True
  \EndIf
  \If{Reset Episode}
    \State  Generate randomized endpoint \& obstacles
	\State Reset robot to initial state and position
	\State Reset time to zero
	\State Start new simulation
  \EndIf
  \State Feed state, action, reward to RL algorithm
 \EndWhile
\end{algorithmic}
\end{algorithm}
\end{figure}

If the rover hit an obstacle, fell off the map, or ran out of time, a penalty was applied and the episode reset. For the former 2, the penalty was 100 and for the latter 20. As the rover made progress towards the goal, the agent received a reward of 100 for each meter, or a penalty of 100 for regression for each meter. 

\begin{equation}
        R_t = C_{veloc}X_t - C_{crash} - C_{fall} - C_{timeout} 
\end{equation}

Training was conducted for 6 days on a desktop workstation, with Intel Core i9-9900K CPU @3.6GHZ, 64GB RAM and a GeForce RTX 2080TI GPU. The simulation was significantly slower than if no only state data was used for training. Around 1270 episodes were completed during this time for the CNN-LSTM + preprocess network setup, with the max time of any single episode being $100 seconds$. Multiple neural network architectures were run at once, with the CPU being the main resource bottleneck restricting the number of architectures trained at a given time.

\subsection{Preprocessing}
\begin{figure}[t]
\begin{center}
   \includegraphics[width=1\linewidth]{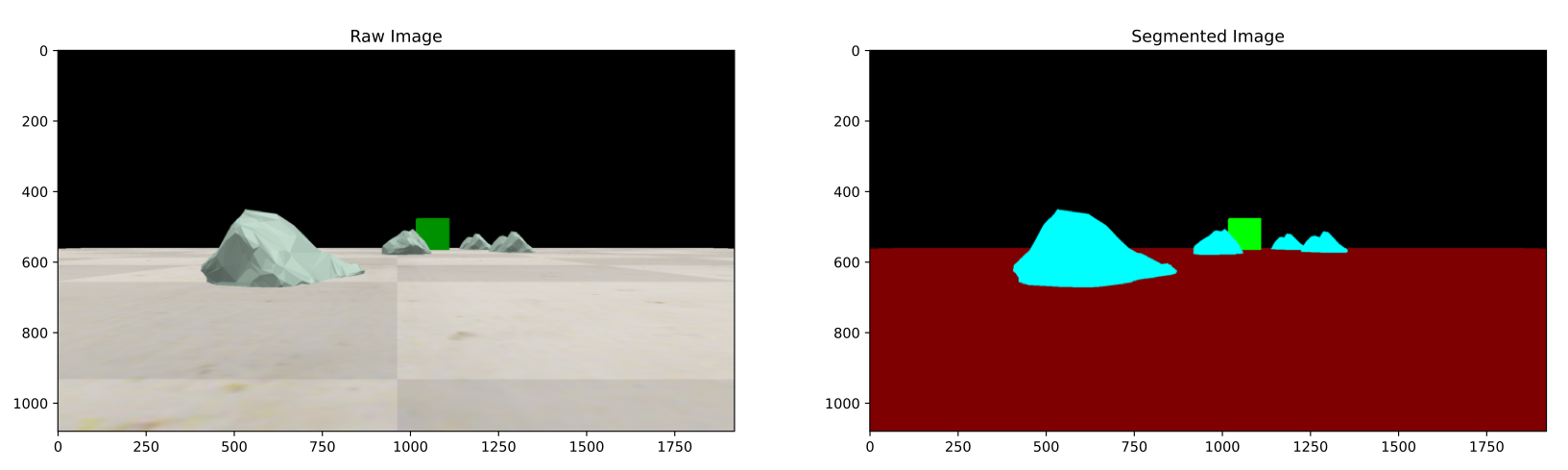}
   \includegraphics[width=1\linewidth]{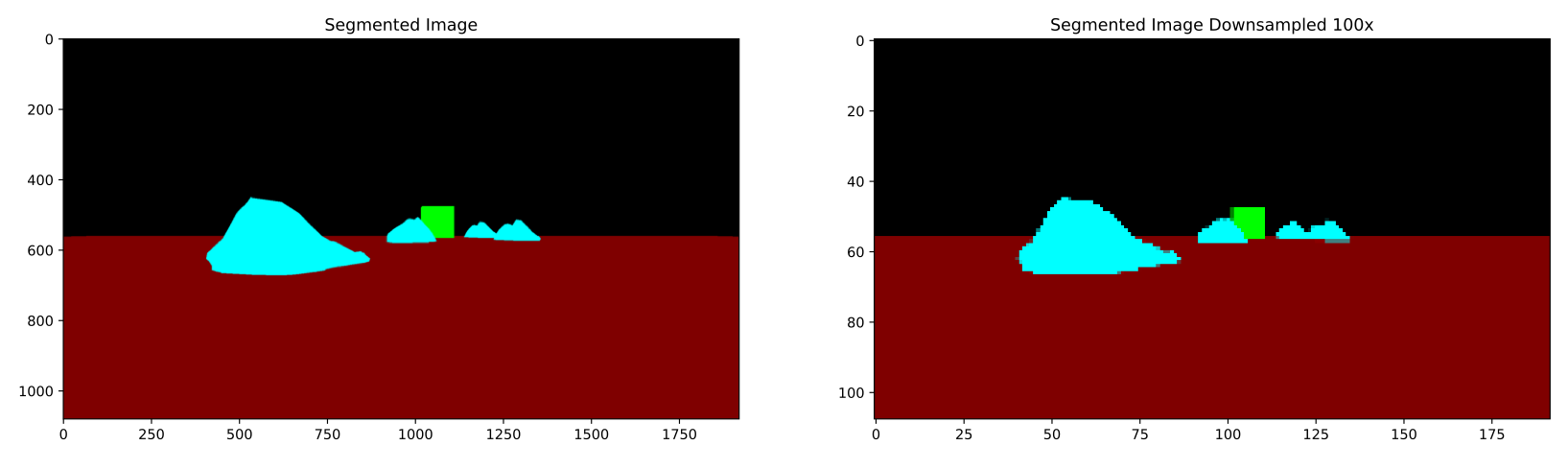}
\end{center}
\caption{The preprocessing is done in two steps: the first one converts the raw image into a semantically segmented image with 4 classes. The second step downsamples it}
\label{fig:preprocessingsteps}
\end{figure}

\begin{figure*}
\begin{center}
\includegraphics[width=\textwidth,height=5cm]{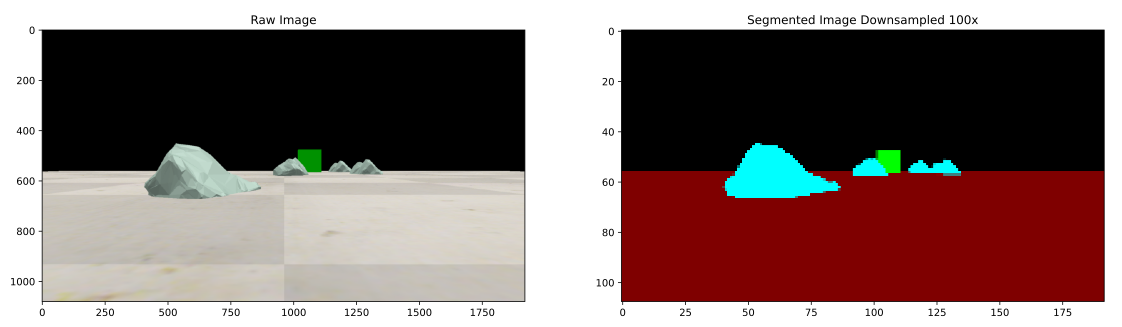}\hspace{0pt}
\caption{An example of a raw image (left) and a preprocessed image (right). }
\label{fig:preprocessing}
\end{center}
\end{figure*}

For this work, we utilize a preprocessing methodology that consists of two steps: first segment the image and second to downsample the image. The broken down steps can be seen in Figure \ref{fig:preprocessingsteps} in the first two rows, with the third row being the resulting product and the fourth row being an example of an image that is downsampled too much, removing some critical information about obstacle locations. A blownup picture of the raw image and preprocessed image can be seen in Figure \ref{fig:preprocessing}.

The first step breaks the image into 4 classes: ground, rock, goal and space. In this way, it is easier to differentiate between safe and unsafe places to drive. The second step reduces the number of pixels by 100 times, from the raw resolution of (1920, 1080, 3) to (192, 108, 3).

\begin{figure}
\begin{center}
   \includegraphics[width=1\linewidth]{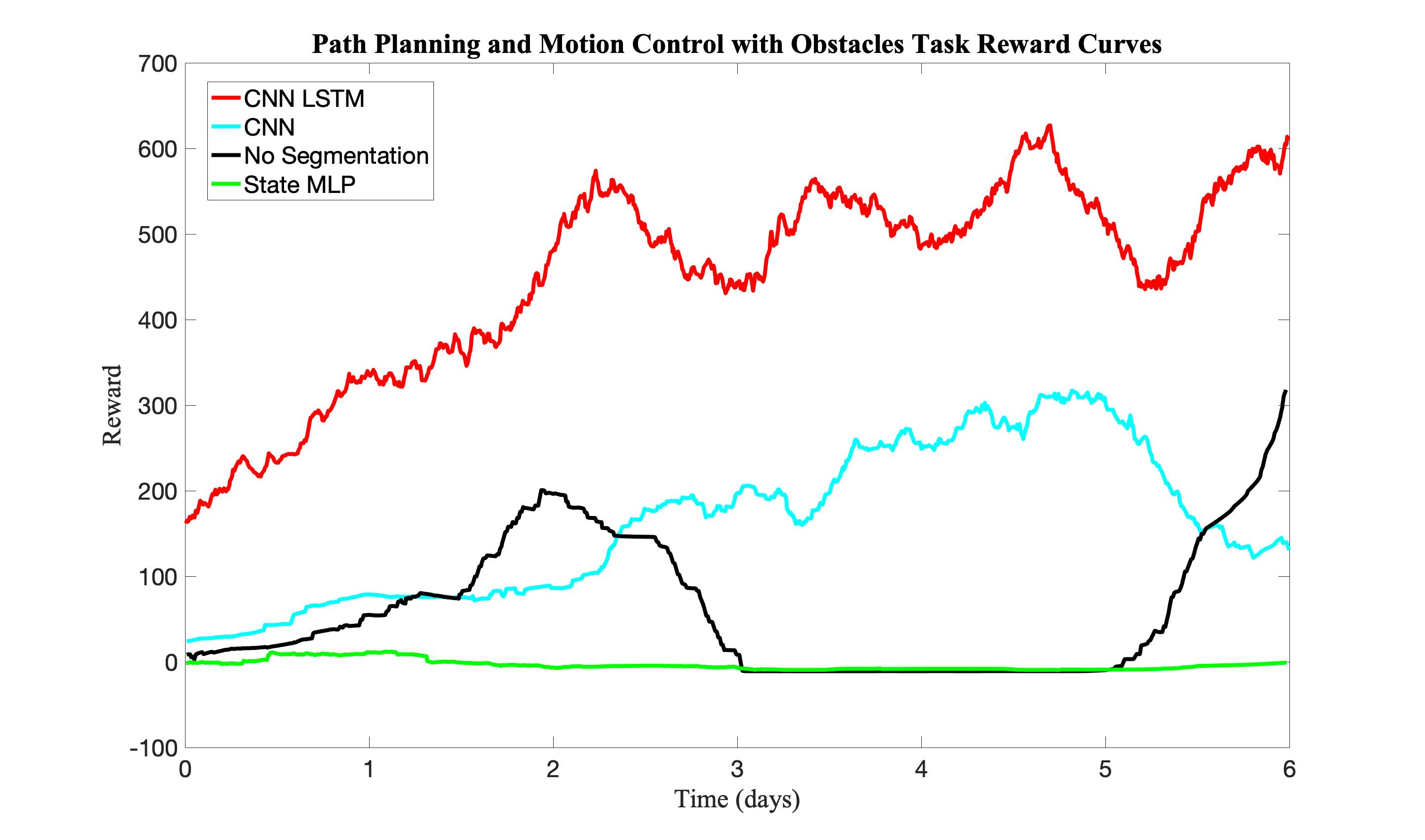}
\end{center}
   \caption{Reward curve shown for four different neural network and preprocessing setups.}
\label{fig:reward}
\end{figure}

\begin{figure}[t]
\begin{center}
   \includegraphics[width=1\linewidth]{SegDS100x.png}
   \includegraphics[width=1\linewidth]{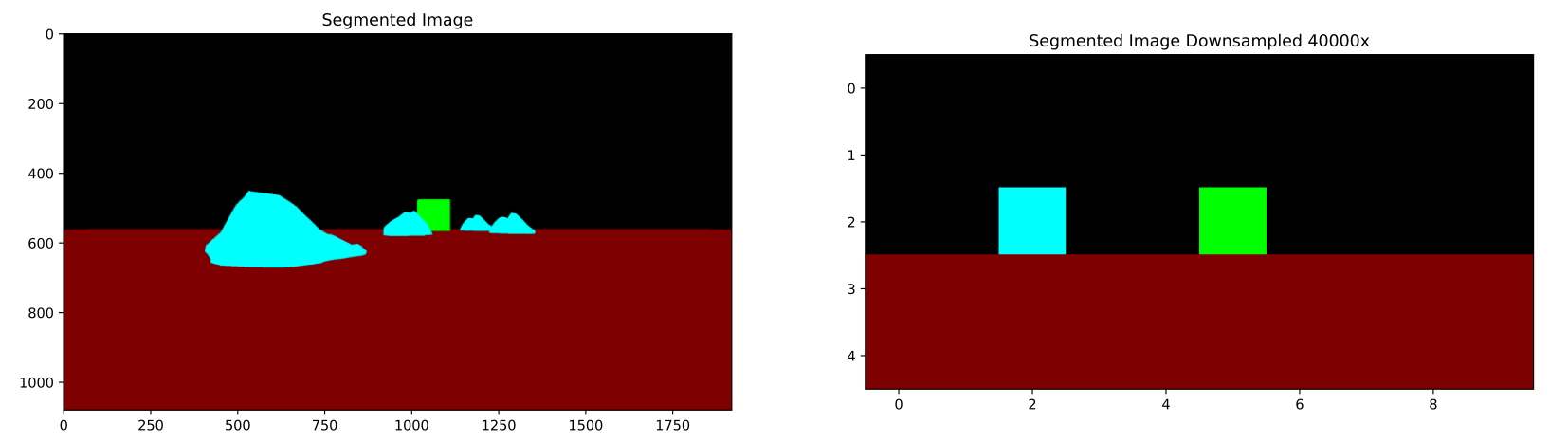}
\end{center}
\caption{An example of a downsampling that does not remove important data (top right) and of an excessively downsampled image that loses information relating to obstacle locations (bottom right). The left side images show the full resolution semantic segmentation image.}
\label{fig:degrading}
\end{figure}

Preprocessing serves three main purposes:
\begin{itemize}
    \item Reducing complexity to make learning easier
    \item Reducing dimensionality to increase computation efficiency 
    \item Reducing the sim2real gap
\end{itemize}

The first two items are shown in the comparison between the CNN-LSTM preprocessed and raw image models. The third item is left for future work. 

For downsampling, Bicubic Interpolation was used. This algorithm works on a 4x4 pixel area and was used as it is known for producing clear images \cite{trusov_analysis_2020}. 

\begin{equation}
  p(x,y) = \sum_{i=0}^3 \sum_{j=0}^3 a_{ij}x^iy^j
\end{equation}

\subsection{Neural Networks and RL algorithm Tuning}
Four neural network architectures were tested for this work. Three of these were based of the Impala Neural Network architecture\cite{espeholt_impala_2018}, consisting of a CNN-LSTM and CNN networks using preprocessed data and a CNN-LSTM network using raw data. Within these networks, the CNN-LSTM that used preprocessed data showed the fastest learning and highest stable reward generated, with the highest success rate. However, the CNN network that used preprocessed data was the only network not to collide with obstacles at all. 

A fourth was a simple MLP network used as a baseline. This baseline was given state data about the robot, such as the angular rotation and velocity, and goal location instead of visual data, however, failed to learn, likely due to obstacle collisions providing too high a disincentive. 

Note that the reward curves are smoothened, accounting for the non-zero start of the CNN LSTM network.

\begin{figure*}
\begin{center}
\includegraphics[width=\textwidth,height=10cm]{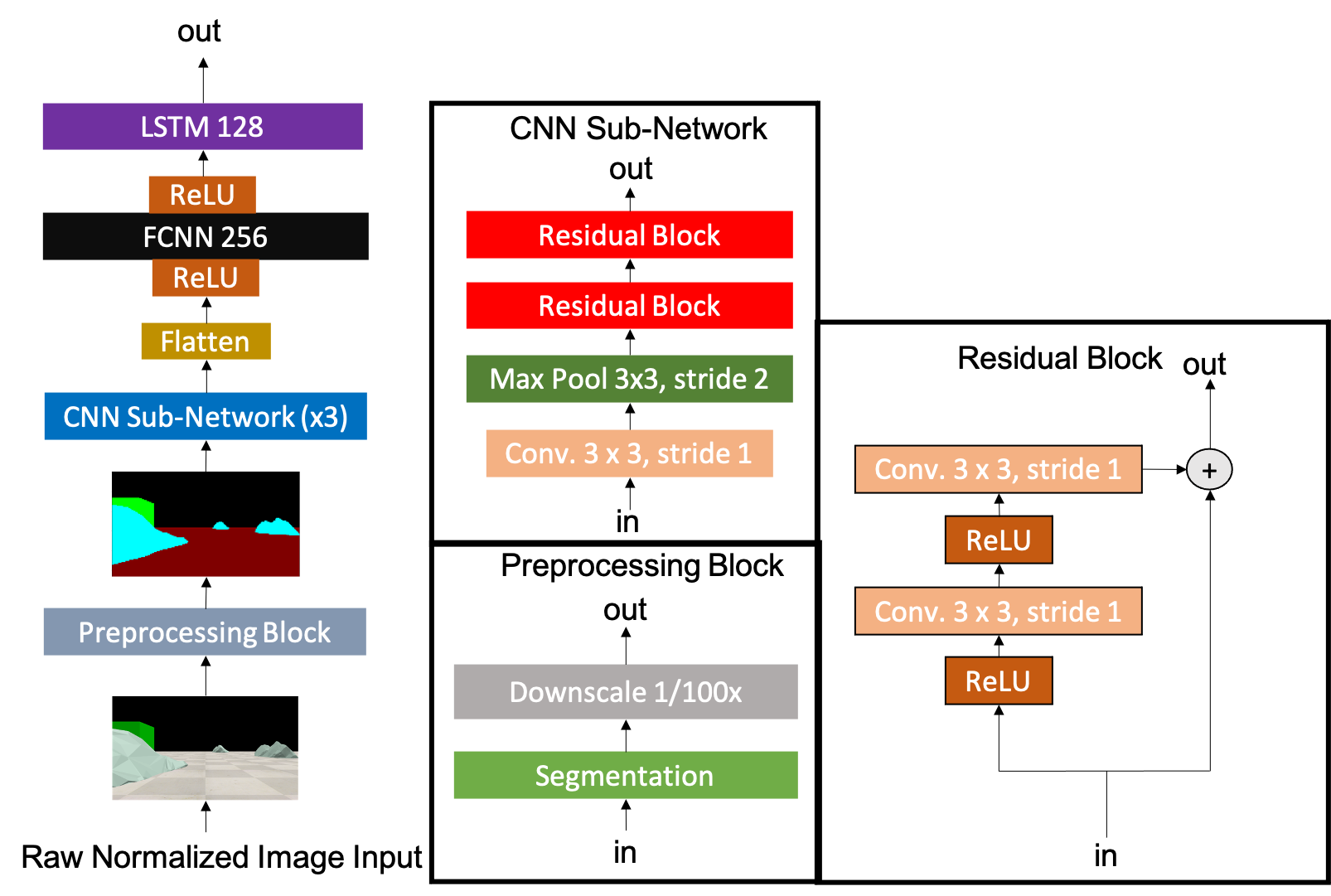}\hspace{0pt}
\caption{\label{NNarch}A CNN-LSTM neural network architecture for containing the visuomotor system. The CNN neural network architecture is the same without the final LSTM layer.}
\label{imagesegmentation}
\end{center}
\end{figure*}

\begin{table}
\caption{Tuned PPO Parameters Used for Training}
\centering
\label{ppoparams}
\begin{tabular}{|l |c |}
\hline
\bf{Parameter}, \bf{Notation} & \bf{Value} \\
\hline
Total Timesteps, $N_{timesteps}$ & 5e6 \\
Learning Rate, $\epsilon$ & 0.0003 \\
Discount Factor, $\gamma$ & 0.85 \\
Entropy Coefficient, $Ent_{coef}$ & 0.01 \\
Epochs per Update, $N_{epochs}$ & 4 \\
Clipping Range, $Clip$ & 0.2 \\
Steps per Update, $N_{step}$ & 64 \\
Advantage Estimation Discount Factor, $\lambda$ & 0.95 \\
Value Function Coefficient, $Vf_{coef}$ & 0.5 \\
Minibatches per Update, $N_{minibatch}$ & 1 \\
Max Grad Norm, $maxG_{norm}$ & 0.5 \\
\hline
\end{tabular}
\end{table}

\subsection{Simulator and Robot}
For our demonstration, we use a simulated robot based off a real robot design. The actual robot was created for swarm applications on the lunar surface, and is small in size with simple skid-steer steering. That is, there are two motors on the robot, one controlling the two left side wheels and one controlling the two right side wheels. For the real rover, the two wheels are connected by a chain, however, in this case we simulate each wheel has an independent motor which might make small variations between real and simulated performance. For skid-steer rovers, turning rate is controlled by the differential between the wheel rotational velocity on the left and right sides, as there is no Ackerman Steering or independent motor to rotate the wheels. This configuration is simpler in design. While the real rover can conduct spot-turning, for this experiment we assume that only positive wheel speeds are possible. The maximum velocity of the rover is set as 0.2m/s, relatively fast given traditional extra-terrestrial rovers such as Curiosity or Perseverance. 

CoppeliaSim is used as it is common in the robotics and space exploration research communities. It was formerly known as V-REP\cite{rohmer_v-rep_2013}. Within CopelliaSim, the physics engine and timestep are definable. For this study, we used the Bullet physics engine\cite{coumans_bulletphysicsbullet3_2013} with a timestep of 0.2 seconds. The rover decisions are updated per 5 simulation timesteps, for a total effective timestep of 1 second between subsequent actions and observations. 

The camera was simulated in CoppeliaSim based off a RealSense D435, which is equipped on the reference rover. This camera has a resolution of (1920, 1080, 3), with a field of view of $69.4\deg$. We set the near/far clipping planes to be 0.01m and 20m respectively. The images were rendered in OpenGL and with perspective mode, a setting which makes the pictured objects get closer as they approach the camera, similar to real life. 

\subsection{Baseline Controllers}
Two baseline controllers were created for reference: a P-controller and a random controller. The P-controller had access to the goal location, information that the DRL controller did not have, however, did not have image data. The error was the angle to the goal. 

The P-controller equation can be describes as follows:
\begin{equation}
  P_{left,out} = P_0-K_{p}e(t)
\end{equation}
\begin{equation}
  P_{right,out} = P_0+K_{p}e(t)
\end{equation}

With $K_p$ equal to 1, the $P_0$ equal to 0.8 and $e(t)$ being equal to the angle to the goal.
\begin{equation}
  P_{left,out} = 0.8-e(t)
\end{equation}
\begin{equation}
  P_{right,out} = 0.8+e(t)
\end{equation}

The motor speeds for the left side were set to $P_{left,out}$ and the motor speeds for the right side were set to $P{right,out}$. This allowed the rover to constantly adjust its heading so that it is driving towards the goal.

The random controller randomly sampled actions. Both these controllers do not adopt to the environment and do not have access to image data.

\section{Results and Discussion}
Using DRL, we were able to develop a visuomotor system that allowed the rover to identify the goal and traverse there while avoiding obstacles. 

We tested 3 configurations, two of these used a preprocessing procedure that transformed raw footage to semantically segmented footage, downscaled it, and then passed it to the DRL agent who processed it using a CNN-LSTM architecture and a CNN architecture, respectively. The last configuration only downscaled the raw footage and used a CNN-LSTM architecture. DRL Config1 refers preprocessing with a CNN-LSTM network. DRL Config2 refers to preprocessing with a CNN network. DRL Config3 refers to raw images with downscaling, and a CNN-LSTM Network. 

We found that the DRL CNN-LSTM architecture with preprocessing learned the fastest within the given 6 days of training. Accordingly, it performed the best on the experiments we conducted, where it achieved 90\% success, no crashes and 10\% timeouts. These results suggest that the agent achieved some ability to dodge obstacles, however, the actual resulting path planning is still sub-optimal. 

Comparatively, the baseline P controller was able to achieve a slightly higher 93.3\% success rate, with 6.7\% crash rate. While this success rate is higher, the a crash is much more undesirable than a timeout.

While this baseline gives a comparison, it should be noted that the P controller has access to the goal location but is blind about the obstacles, whereas the DRL controllers have no a-priori knowledge of either the goal location or the obstacles, and have to infer it all via the image. Furthermore, the problem is partially observable, and in certain circumstances the rover could lose sight of the goal location, or it could be obscured behind a rock obstacle. In this way, the rover needs to perform either an exploration or perhaps a memory recall using the LSTM structure. 

Although the CNN-LSTM network was able to score a higher total reward, the CNN network was able to better avoid collisions with obstacles. Both these networks had a higher success rate than the reactive and random baseline controllers used. Non-success trials were classified into collision, fall and timeout. 

The results are significant for the following reasons:
\begin{itemize}
    \item the development of a visuomotor system is possible by mixing CNNs and RL with access to vision data 
    \item the NN architecture can have a significant impact on the performance of the visuomotor system
    \item preprocessing the data can make learning easier
    \item DRL controllers are able to adapt to the environment conditions, such as frequency of actions, whereas the baseline controllers cannot
\end{itemize}

\begin{table}
\begin{center}
\begin{tabular}{|l|c|c|c|c|}
\hline
Method & Success & Collision & Fall & Timeout \\
\hline\hline
DRL Config1 & 90\% & 0\% & 0\% & 10\% \\
DRL Config2 & 53.3\%  & 0\% & 6.7\% & 40\%\\
DRL Config3 & 60\% & 13.3\% & 0\% & 26.7\%\\
P Controller & 93.3\%  & 6.67\% & 0\% & 0\%\\
\hline
\end{tabular}
\end{center}
\caption{Results from 30 trial test runs similar to the training conditions. DRL Config 1 and the P Controller has similar success rate however, the P controller had collisions while the DRL controller only had timeouts.}
\end{table}

\section{Conclusion and Future Work}
This work demonstrated the development of a visuomotor system that allowed a robotic agent to reach the goal while avoiding obstacles with a 90\% success rate, reaching a comparative scores to the P-controller. While the P-controller crashed into obstacles, the DRL controller was able to avoid the obstacles. However, the DRL controller path was suboptimal and so in future work, we should find ways to increase the directness of the path taken.

With the current setup it was hard to a significant performance difference between the DRL controller and the P-controller by looking only at the success rate. For future work, we will increase the number of obstacles, so that during training the agent can have more experience avoiding them, and so that during a demonstration, the success rate difference between the P-controller and the DRL controller can be highlighted. 

One conceivable large advantage of preprocessing the neural network is that a similar preprocessing procedure could be done in both simulation and the real world to simplify the sim2real gap. We leave it for future work to verify this. 

Future work includes training for longer and under different conditions to try to achieve stable learning through no segmentation and non-LSTM networks, as well as higher learning under CNN LSTM structure. 

{\small
\bibliographystyle{ieee_fullname}
\bibliography{visuomotor.bib}
}

\end{document}